\pdfoutput=1

\documentclass[11pt]{article}

\usepackage[]{ACL2023}

\usepackage{times}
\usepackage{latexsym}
\usepackage{booktabs}
\usepackage{setspace}
\usepackage[T1]{fontenc}

\usepackage[utf8]{inputenc}

\usepackage{microtype}

\usepackage{inconsolata}

\usepackage{amsmath}
\usepackage{amssymb}
\usepackage{mathtools}
\usepackage{amsthm}

\usepackage[capitalize,noabbrev]{cleveref}

\theoremstyle{plain}

\usepackage[textsize=tiny]{todonotes}

\usepackage{amsmath, amssymb}
\usepackage{xcolor}
\usepackage{graphicx}
\usepackage{soul}
\usepackage{algorithm}
\usepackage{multirow}
\usepackage{array}
\usepackage{xcolor, colortbl}
\usepackage{graphicx}
\newcolumntype{P}[1]{>{\centering\arraybackslash}p{#1}}

\definecolor{yy}{RGB}{0,120,120}

\title{In and Out-of-Domain Text Adversarial Robustness via Label Smoothing}

\author{Yahan Yang\Thanks{ The first two authors contributed equally to this paper. Most of the work done while Soham Dan was at the University of Pennsylvania.}  \\ University of Pennsylvania \\  \texttt{yangy96@seas.upenn.edu}
         \And
        Soham Dan\samethanks\\ IBM Research \\ \texttt{soham.dan@ibm.com} \AND
        Dan Roth \\ University of Pennsylvania \\ \texttt{danroth@seas.upenn.edu}  \And 
        Insup Lee \\ University of Pennsylvania \\ \texttt{lee@cis.upenn.edu}}

\begin{document}
\maketitle
\begin{abstract}
Recently it has been shown that state-of-the-art NLP models are vulnerable to adversarial attacks, where the predictions of a model can be drastically altered by slight modifications to the input (such as synonym substitutions). While several defense techniques have been proposed, and adapted, to the discrete nature of text adversarial attacks, the benefits of general-purpose regularization methods such as label smoothing for language models, have not been studied. In this paper, we study the adversarial robustness provided by label smoothing strategies in foundational models for diverse NLP tasks in both in-domain and out-of-domain settings. Our experiments show that label smoothing significantly improves adversarial robustness in pre-trained models like BERT, against various popular attacks. We also analyze the relationship between prediction confidence and robustness, showing that label smoothing reduces over-confident errors on adversarial examples. 
\end{abstract}

\section{Introduction}
Neural networks are vulnerable to adversarial attacks: small perturbations to the input ,which do not fool humans \cite{szegedy2013intriguing,goodfellow2014explaining,madry2017towards}. In NLP tasks, previous studies \cite{alzantot2018generating,textfooler,bert-attack,bae} demonstrate that simple word-level text attacks (synonym substitution, word insertion/deletion) easily fool state-of-the-art models, including pre-trained transformers like BERT \cite{bert, wolf-etal-2020-transformers}. Further, it has recently been shown models are overconfident\footnote{Confidence on an example is the highest softmax score of the classifier prediction on that example.} on examples which are easy to attack \cite{qin2021improving} and indeed, such over-confident predictions plague much of modern deep learning \cite{kong2020calibrated, guo2017calibration, nguyen2015deep, rahimi2020intra}. Label smoothing is a regularization method that has been proven effective in a variety of applications, and modalities \cite{szegedy2016rethinking, chorowski2017towards, vaswani2017attention}. Importantly, it has been shown to reduce overconfident predictions and produce better confidence calibrated classifiers \cite{muller2019does,zhang2021delving,DanRo21,desai2020calibration, huang2021context,liu2020class}. 

In this work, we focus on the question: \textit{does label smoothing also implicitly help in adversarial robustness?} While there has been some investigation in this direction for adversarial attacks in computer vision, \cite{fu2020label,goibert2019adversarial,shafahi2019label}, there is a gap in understanding of whether it helps with discrete, text adversarial attacks used against NLP systems. 
With the increasing need for robust NLP models in safety-critical applications and a lack of generic robustness strategies,\footnote{which are flexible, simple and not over-specialized to very specific kinds of text adversarial attacks.} there is a need to understand inherent robustness properties of popular label smoothing strategies, and the interplay between confidence and robustness of a model. 

In this paper, we extensively study standard label smoothing and its adversarial variant, covering robustness, prediction confidence, and domain transfer properties. We observe that label smoothing provides implicit robustness against adversarial examples. Particularly, we focus on pre-trained transformer models and test robustness under various kinds of black-box and white-box word-level adversarial attacks, in both in-domain and out-of-domain scenarios. 
Our experiments show that label smoothing (1) improves robustness to text adversarial attacks (both black-box and white-box), and (2) mitigates over-confident errors on adversarial textual examples. Analysing the adversarial examples along various quality dimensions reveals the remarkable efficacy of label smoothing as a simple add-on robustness and calibration tool.

\section{Background}

\subsection{Text Adversarial Attacks}
Our experiments evaluate the robustness of text classification models under three state-of-the-art text adversarial attacks TextFooler (black-box), BAE (black-box) and SemAttack (white-box), described below.\footnote{The black-box attacks keep querying the model with its attempts until the victim model is fooled while the white-box attack has access to the gradients to the model. Further details of the attacks are in \cite{textfooler,bae,semattack}.}For a particular victim NLP model and a raw text input, the attack produces semantically-similar adversarial text as output. Importantly, only those examples are attacked, which are originally correctly predicted by the victim model. The attacks considered are word-level, i.e. they replace words in a clean text with their synonyms to maintain the meaning of the clean text, but change the prediction of the victim models. 
\vspace{-0.25em}
\begin{itemize}
\itemsep0em 
    \item \textbf{TextFooler (TF)}: \cite{textfooler} proposes an attack which determines the word importance in a sentence, and then replaces the important words with qualified synonyms. 
    \item \textbf{BAE}: \cite{bae} uses masked pre-trained language models to generate replacements for the important words until the victim model's prediction is incorrect.  
    \item \textbf{SemAttack (SemAtt)}: \citep{semattack} introduces an attack to search perturbations in the contextualized embedding space by formulating an optimization problem as in \citep{cw2}. We specifically use the white-box word-level version of this attack.
\end{itemize}
 \vspace{-1em}

\subsection{Label Smoothing}
Label Smoothing is a modified fine-tuning procedure to address overconfident predictions. It introduces uncertainty to smoothen the posterior distribution over the target labels. Label smoothing has been shown to implicitly calibrate neural networks on out-of-distribution data, where \textit{calibration} measures how well the model confidences are aligned with the empirical likelihoods \cite{guo2017calibration}.
\vspace{-0.25em}
\begin{itemize}
\itemsep0em 
    \item \textbf{Standard Label Smoothing (LS)} \cite{szegedy2013intriguing,muller2019does} constructs a new
target vector ($y_i^{LS}$) from the one-hot target vector ($y_i$), where $y_i^{LS} = (1-\alpha)y_i+\alpha/K$ for a $K$ class classification problem. $\alpha$ is a hyperparameter selection and its range is from 0 to 1. 

    \item \textbf{Adversarial Label Smoothing (ALS)} \cite{goibert2019adversarial} constructs a new target vector ($y_i^{ALS}$) with a
probability of $1-\alpha$ on the target label and $\alpha$ on the label to which the classification model assigns the minimum softmax scores, thus introducing uncertainty. 
\end{itemize}
\vspace{-0.25em}
 For both LS and ALS, the cross entropy loss is subsequently minimized between the model predictions and the modified target vectors $y_i^{LS},y_i^{ALS}$.
\vspace{-0.5em}
%

%

\vspace{-1.0em}
\section{Experiments}
\vspace{-0.2em}
In this section, we present a thorough empirical evaluation on the effect of label smoothing on adversarial robustness for two pre-trained transformer models: BERT and its distilled variant, dBERT, which are the victim models. \footnote{Additional results on more datasets, models, other attacks and $\alpha$ values, are presented in the Appendix. } 
We attack the victim models using TF, BAE, and SemAttack. 
For each attack, we present results on both the standard models and the label-smoothed models on various classification tasks: text classification and natural language inference. For each dataset we evaluate on a randomly sampled subset of the test set (1000 examples), as done in prior work \cite{clare,textfooler,bae}. 
 We evaluate on the following tasks, and other details about the setting is in Appendix \ref{exp}:

 \begin{table*}[]
 \singlespacing
\footnotesize
\begin{tabular}{P{3em} P{2em} P{2em}P{2em} P{2em}P{2em}P{2em}} \toprule
\textbf{SST-2}   & \multicolumn{2}{c}{\begin{tabular}{@{}c@{}}Clean \\ Acc ($\uparrow$) \end{tabular}}  & \multicolumn{2}{c}{\begin{tabular}{@{}c@{}} Attack Success \\ Rate ($\downarrow$) \end{tabular}} &\multicolumn{2}{c}{\begin{tabular}{@{}c@{}}Adv \\ Conf ($\downarrow$) \end{tabular}} \\ \hline
BERT($\alpha$) & 0 & 0.45 & 0 & 0.45 & 0 & 0.45\\ \hline
TF  & 91.97 & \textbf{92.09}  & 96.38 & \textbf{88.92} & 78.43 & \textbf{63.62}  \\
BAE  & 91.97 &\textbf{92.09} & 57.11 & \textbf{53.42}  & 86.92  & \textbf{68.35}  \\ 
SemAtt  & 91.97 & \textbf{92.09}& 86.41 & \textbf{54.05}  &  80.12 &  \textbf{64.55} \\  \hline
dBERT($\alpha$) & 0 & 0.45 & 0 & 0.45 & 0 & 0.45\\ \hline
TF  & 89.56 &  \textbf{89.68}  & 96.29 & \textbf{89.77} & 76.28 & \textbf{61.60} \\
BAE  & 89.56 &  \textbf{89.68} & 59.28  & \textbf{56.52} & 83.55 & \textbf{66.11}   \\ 
SemAtt & 89.56 &  \textbf{89.68} & 91.68 & \textbf{69.69}  &  78.93 & \textbf{62.42}  \\  
\end{tabular}
\begin{tabular}{P{3em}P{2em}P{2em}P{2em}P{2em}P{2em}P{2em}} \toprule
\textbf{AG\_news}   & \multicolumn{2}{c}{\begin{tabular}{@{}c@{}}Clean \\ Acc ($\uparrow$) \end{tabular}}  & \multicolumn{2}{c}{\begin{tabular}{@{}c@{}} Attack Success \\ Rate ($\downarrow$) \end{tabular}} &\multicolumn{2}{c}{\begin{tabular}{@{}c@{}}Adv \\ Conf ($\downarrow$) \end{tabular}} \\ \hline
BERT($\alpha$) & 0 & 0.45 & 0 & 0.45 & 0 & 0.45\\ \hline
TF & \textbf{94.83} & 94.67 & 88.26 & \textbf{77.47} & 59.02 & \textbf{42.46} \\
BAE  & \textbf{94.83} & 94.67 & 74.83 & \textbf{62.82} & 60.66 & \textbf{43.98}\\ 
SemAtt & \textbf{94.83} & 94.67 & 52.65 & \textbf{30.49}  &  62.32 & \textbf{44.99}  \\  \hline
dBERT($\alpha$) & 0 & 0.45 & 0 & 0.45 & 0 & 0.45\\ \hline
TF  & \textbf{94.73} & 94.47 & 90.11 & \textbf{74.52} & 57.60 &  \textbf{41.40}    \\
BAE & \textbf{94.73} & 94.47    & 77.79 & \textbf{63.65} & 60.01 & \textbf{42.74} \\
SemAtt & \textbf{94.73} & 94.47  & 52.07 & \textbf{34.05}  &  60.40 & \textbf{43.27}  \\  
\end{tabular}
\begin{tabular}{P{3em} P{2em} P{2em}P{2em} P{2em}P{2em}P{2em}} \toprule
\textbf{Yelp}& \multicolumn{2}{c}{\begin{tabular}{@{}c@{}}Clean \\ Acc ($\uparrow$) \end{tabular}}  & \multicolumn{2}{c}{\begin{tabular}{@{}c@{}} Attack Success \\ Rate ($\downarrow$) \end{tabular}} &\multicolumn{2}{c}{\begin{tabular}{@{}c@{}}Adv \\ Conf ($\downarrow$) \end{tabular}} \\ \hline
BERT($\alpha$) & 0 & 0.45 & 0 & 0.45 & 0 & 0.45\\ \hline
TF  & \textbf{97.73} & 97.7 & 99.32 & \textbf{92.90}  & 64.85  & \textbf{55.36} \\
BAE  & \textbf{97.73} & 97.7 & 55.35 & \textbf{45.14}  &  68.28  & \textbf{57.38}   \\  
SemAtt  & \textbf{97.73} & 97.7 & 93.55 & \textbf{36.17}  & 74.53  & \textbf{60.24} \\  \hline
dBERT($\alpha$) & 0 & 0.45 & 0 & 0.45 & 0 & 0.45\\ \hline
TF & \textbf{97.47} & 97.4 & 99.45 & \textbf{93.36} & 61.75 & \textbf{54.63} \\
BAE  & \textbf{97.47} & 97.4 & 58.14 & \textbf{45.59} & 64.27 & \textbf{57.14} \\ 
SemAtt  & \textbf{97.47} & 97.4 & 97.37 & \textbf{43.92} & 71.34 & \textbf{60.57}  \\  
\bottomrule
\end{tabular}
\begin{tabular}{P{3em} P{2em} P{2em}P{2em} P{2em}P{2em}P{2em}} \toprule

\textbf{SNLI}  & \multicolumn{2}{c}{\begin{tabular}{@{}c@{}}Clean \\ Acc ($\uparrow$) \end{tabular}}  & \multicolumn{2}{c}{\begin{tabular}{@{}c@{}} Attack Success \\ Rate ($\downarrow$) \end{tabular}} &\multicolumn{2}{c}{\begin{tabular}{@{}c@{}}Adv \\ Conf ($\downarrow$) \end{tabular}} \\ \hline
BERT($\alpha$) & 0 & 0.45 & 0 & 0.45 & 0 & 0.45\\ \hline
TF & \textbf{89.56} & 89.23 & 96.5 & \textbf{96.15} & 68.27 & \textbf{52.61}\\
BAE & \textbf{89.56} & 89.23 & 74.95 & \textbf{74.82} & 76.13 & \textbf{57.42} \\ 
SemAtt & \textbf{89.56} & 89.23 & 99.11 & \textbf{91.94} & 75.41 & \textbf{58.01} \\ \hline
dBERT($\alpha$) & 0 & 0.45 & 0 & 0.45 & 0 & 0.45\\ \hline
TF  & \textbf{87.27} & 87.1 & 98.12 & \textbf{96.86} & 65.19 & \textbf{50.80} \\
BAE  & \textbf{87.27} & 87.1 & 74.08 & \textbf{72.91} & 72.89  & \textbf{55.49}  \\ 
SemAtt & \textbf{87.27} & 87.1 & 98.43 & \textbf{92.84} & 71.17 & \textbf{54.96} \\ 
\bottomrule
\end{tabular}
\vspace{-1.0em}
\caption{Comparison of standard models and models fine-tuned with standard label smoothing techniques (LS) against various attacks for in-domain data. We show clean accuracy, attack success rate and average confidence on successful adversarial texts. For each dataset, the left column are the results for standard model, and the right column are for LS models where $\alpha$ denotes the label smoothing factor ($\alpha$=0: no LS). $\uparrow$ ($\downarrow$) denotes higher (lower) is better respectively. dBERT denotes the distilBERT model.}
\label{main}
\vspace{-1.5em}
\end{table*}
\vspace{-0.5em}
\begin{itemize}
\itemsep0em 
\item \textbf{Text Classification}: We evaluate on movie review classification using Movie Review (MR) \cite{mrdataset} and Stanford Sentiment Treebank (SST2) \cite{socher2013recursive} (both binary classification), restaurant review classification: Yelp Review \cite{yelpdataset} (binary classification), and news category classification: AG News \cite{agnews} (having the following four classes: World, Sports, Business, Sci/Tech). 
\item \textbf{Natural Language Inference:} We investigate two datasets for this task: the Stanford Natural Language Inference Corpus (SNLI) \cite{snli} and the Multi-Genre Natural Language Inference corpus (MNLI) \cite{mnli}, both having three classes. For MNLI, our work only evaluates performance on the matched genre test-set in the OOD setting presented in subsection \ref{ood} . 
\end{itemize}

\vspace{-1em}
\subsection{In-domain Setting}
\label{sec:in_domain}

In the in-domain setting (iD), the pre-trained transformer models are fine-tuned on the train-set for each task and evaluated on the corresponding test-set. For each case, we report the clean accuracy, the adversarial attack success rate (percentage of misclassified examples after an attack) and the average confidence on successfully attacked examples (on which the model makes a wrong prediction).\footnote{Details of each metric are presented in Appendix \ref{eval_metric}.} Table \ref{main} shows the performance of BERT and dBERT, with and without label-smoothing.  We choose label smoothing factor $\alpha = 0.45$ for standard label-smoothed models in our experiments. 

\begin{figure}[!htb]
\centering
\vspace{-0.1em}
\includegraphics[width=0.75\linewidth]{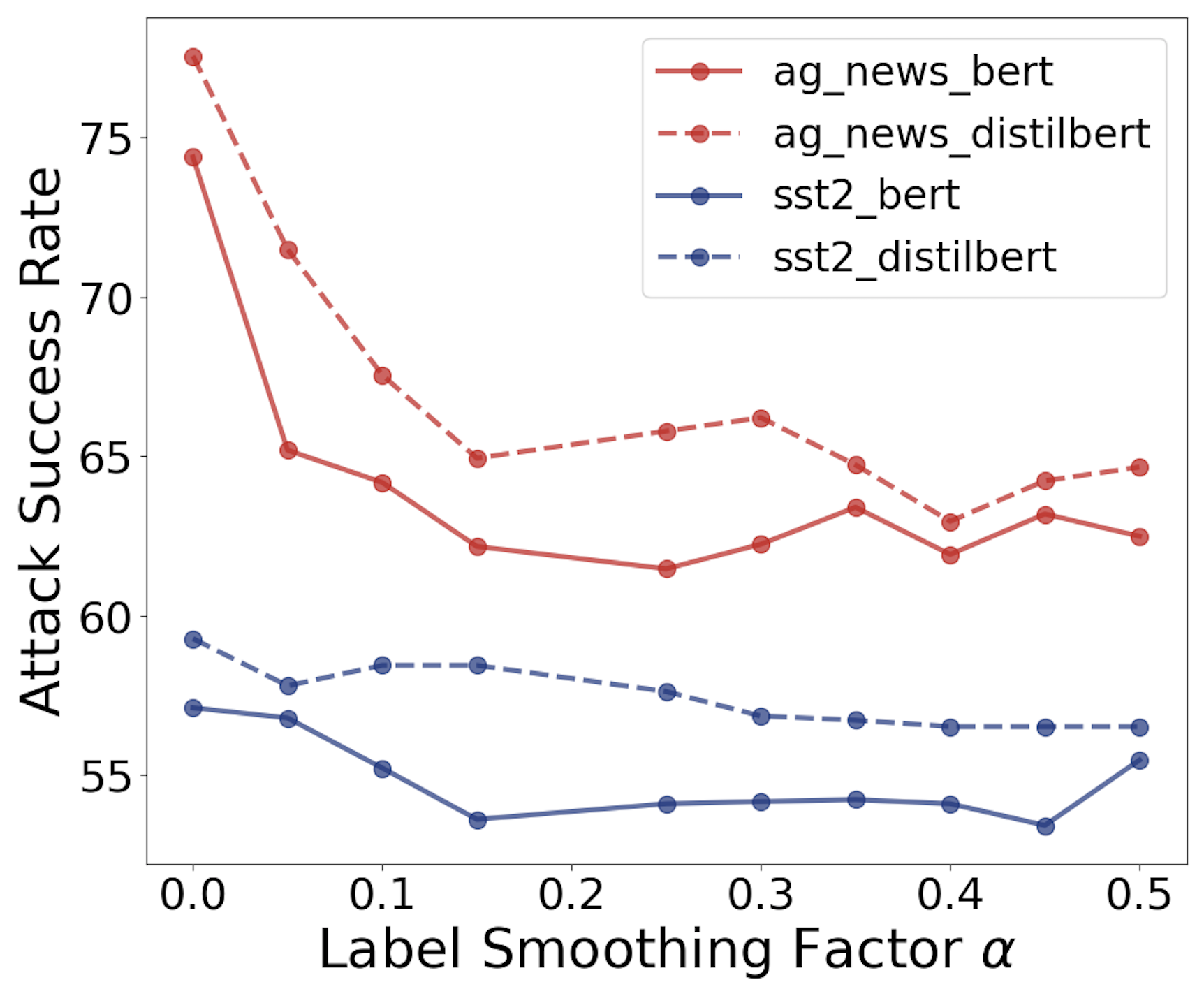}
\vspace{-0.5em}
\caption{Adversarial success rate versus label smoothing factors (on AG News and SST-2 with BAE attack.) }
\label{fig:alpha_vs_succ}
\vspace{-1em}
\end{figure}

We see that label-smoothed models are more robust for every adversarial attack across different datasets in terms of the attack success rate, which is a standard metric in this area \citep{clare, queryattack2022}.  Additionally, the higher confidence of the standard models on the successfully attacked examples indicates that label smoothing helps mitigate overconfident mistakes in the adversarial setting. Importantly, the clean accuracy remains almost unchanged in all the cases. Moreover, we observe that the models gain much more robustness from LS under white-box attack, compared to the black-box setting. We perform hyperparameter sweeping for the label smoothing factor $\alpha$ to investigate their impact to model accuracy and adversarial robustness. Figure \ref{fig:alpha_vs_succ} shows that the attack success rate gets lower as we increase the label smooth factor when fine-tuning the model while the test accuracy is comparable\footnote{More results for different $\alpha$ values are in Appendix \ref{sec:alpha_sweep_textfooler}}. However, when the label smoothing factor is larger than $0.45$, there is no further improvement on adversarial robustness in terms of attack success rate. Automatic search for an optimal label smoothing factor and its theoretical analysis is important future work. 


\begin{table}[]
\footnotesize
 \singlespacing
\vspace{-1em}
\begin{tabular}{P{3em} P{2em} P{2em}P{2em} P{2em}P{2em}P{2em}} \toprule
\textbf{SNLI}   & \multicolumn{2}{c}{\begin{tabular}{@{}c@{}}Clean \\ Acc ($\uparrow$) \end{tabular}}  & \multicolumn{2}{c}{\begin{tabular}{@{}c@{}} Attack Success\\Rate ($\downarrow$)\end{tabular}} &\multicolumn{2}{c}{\begin{tabular}{@{}c@{}}Adv \\ Conf ($\downarrow$) \end{tabular}} \\ \hline
BERT($\alpha$) & 0 & 0.45 & 0 & 0.45 & 0 & 0.45\\ \hline
TF & \textbf{89.56} & 88.5 & 96.5 & 96.5 & 68.27 & \textbf{41.22}\\
BAE & \textbf{89.56} & 88.5 & 74.95 & \textbf{74.87} & 76.13 & \textbf{44.93}\\
SemAtt & \textbf{89.56} & 88.5 & 99.11 & \textbf{91.53} & 75.41 & \textbf{44.97} \\
\bottomrule
\textbf{AG\_news}   & \multicolumn{2}{c}{\begin{tabular}{@{}c@{}}Clean \\ Acc ($\uparrow$) \end{tabular}}  & \multicolumn{2}{c}{\begin{tabular}{@{}c@{}} Attack Success\\Rate ($\downarrow$)\end{tabular}} &\multicolumn{2}{c}{\begin{tabular}{@{}c@{}}Adv \\ Conf ($\downarrow$) \end{tabular}} \\ \hline
BERT($\alpha$) & 0 & 0.45 & 0 & 0.45 & 0 & 0.45\\ \hline
TF & \textbf{94.83} & 94.37 & 88.26 & \textbf{77.74} & 59.02 & \textbf{32.87} \\
BAE  & \textbf{94.83} & 94.37 & 74.83 & \textbf{64.15} & 60.66 & \textbf{33.45} \\ 
SemAtt & \textbf{94.83} & 94.37 & 52.65 & \textbf{27.13} &  62.32 & \textbf{34.72} \\ 
\bottomrule
\end{tabular}
\vspace{-1em}
\caption{Comparison of standard models versus models trained with ALS against various attacks on SNLI and AG\_news.  $\uparrow$ ($\downarrow$) denotes higher (lower) is better respectively.}
\label{advlabel}
\vspace{-2em}
\end{table}

We also investigate the impact of adversarial label smoothing (ALS) and show that the adversarial label smoothed methods also improves model's robustness in Table \ref{advlabel}.

\subsection{Out-of-Domain setting}
\vspace{-0.25em}
\label{ood}
We now evaluate the benefits of label smoothing for robustness in the out-of-domain (OOD) setting, where the pre-trained model is fine-tuned on a particular dataset and is then evaluated directly on a different dataset, which has a matching label space. Three examples of these that we evaluate on are the Movie Reviews to SST-2 transfer, the SST-2 to Yelp transfer, and the SNLI to MNLI transfer.

\begin{table}[!b]
\footnotesize
 \singlespacing
\centering
\vspace{-2.5em}
\begin{tabular}{P{4em} P{1.8em} P{1.8em}P{1.8em} P{1.8em}P{1.8em}P{1.8em}} \toprule
\textbf{MR$\rightarrow$SST2}  & \multicolumn{2}{c}{\begin{tabular}{@{}c@{}}Clean \\ Acc ($\uparrow$) \end{tabular}}  & \multicolumn{2}{c}{\begin{tabular}{@{}c@{}} Attack Success \\ Rate ($\downarrow$) \end{tabular}} &\multicolumn{2}{c}{\begin{tabular}{@{}c@{}}Adv \\ Conf ($\downarrow$) \end{tabular}} \\ \hline
BERT ($\alpha$) & 0 & 0.45 & 0 & 0.45 & 0 & 0.45\\ \hline
TF  & 90.71 & \textbf{91.06} & \textbf{90.9}  & 90.93 & 69.47  & \textbf{58.41}  \\
BAE   & 90.71 & \textbf{91.06} & \textbf{62.83} &  63.1 & 75.2  & \textbf{62.6} \\ 
SemAtt  & 90.71 & \textbf{91.06}  & 82.68 & \textbf{76.07} & 67.64 & \textbf{57.9}  \\\hline
dBERT($\alpha$) & 0 & 0.45 & 0 & 0.45 & 0 & 0.45\\ \hline
TF  &  88.19 & \textbf{88.99}  & \textbf{94.28} & 94.59 & 64.95 & \textbf{57.2}  \\
BAE  &  88.19 & \textbf{88.99}  & \textbf{65.41} & 65.72 & 71.89 & \textbf{61.5}  \\ 
SemAtt  &  88.19 & \textbf{88.99}  & 88.56 & \textbf{86.21} & 66.51 & \textbf{58.14}  \\
\end{tabular}
\begin{tabular}{P{4em} P{1.8em} P{1.8em}P{1.8em} P{1.8em}P{1.8em}P{1.8em}} \toprule
{\begin{tabular}{@{}c@{}}\textbf{SNLI}$\rightarrow$\textbf{MNLI} \end{tabular}}  & \multicolumn{2}{c}{\begin{tabular}{@{}c@{}}Clean \\ Acc ($\uparrow$) \end{tabular}}  & \multicolumn{2}{c}{\begin{tabular}{@{}c@{}} Attack Success \\ Rate ($\downarrow$) \end{tabular}} &\multicolumn{2}{c}{\begin{tabular}{@{}c@{}}Adv \\ Conf ($\downarrow$) \end{tabular}} \\ \hline
BERT ($\alpha$) & 0 & 0.45 & 0 & 0.45 & 0 & 0.45\\ \hline
TF  & \textbf{73.4} & 72.1  & 94.82 & \textbf{92.79}  & 58.04 & \textbf{46.43}   \\
BAE  & \textbf{73.4} & 72.1  & 82.56 &  \textbf{80.72} & 63.00  & \textbf{49.45}  \\ 
SemAtt   & \textbf{73.4} & 72.1 & 99.73 & \textbf{98.75} & 60.32 & \textbf{47.35}  \\
\hline
dBERT($\alpha$) & 0 & 0.45 & 0 & 0.45 & 0 & 0.45\\ \hline
TF  & \textbf{65.4} & 62.1 & 94.50 & \textbf{92.59}    & 54.54 & \textbf{44.81}  \\
BAE  & \textbf{65.4} & 62.1 & 77.68 & \textbf{75.52}  & 58.88  & \textbf{47.83}  \\ 
SemAtt  & \textbf{65.4} & 62.1  & 99.39 & \textbf{96.78} & 57.10 & \textbf{45.43}  \\
\end{tabular}
\begin{tabular}{P{4em} P{1.8em} P{1.8em}P{1.8em} P{1.8em}P{1.8em}P{1.8em}} \toprule
{\begin{tabular}{@{}c@{}}\textbf{SST-2} $\rightarrow$ \textbf{Yelp} \end{tabular}}  & \multicolumn{2}{c}{\begin{tabular}{@{}c@{}}Clean \\ Acc ($\uparrow$) \end{tabular}}  & \multicolumn{2}{c}{\begin{tabular}{@{}c@{}} Attack Success \\ Rate ($\downarrow$) \end{tabular}} &\multicolumn{2}{c}{\begin{tabular}{@{}c@{}}Adv \\ Conf ($\downarrow$) \end{tabular}} \\ \hline
BERT ($\alpha$) & 0 & 0.45 & 0 & 0.45 & 0 & 0.45\\ \hline
TF  & \textbf{92.5} & 92.4   & 99.57 & \textbf{98.27} & 60.80  & \textbf{54.28}   \\
BAE  & \textbf{92.5} & 92.4 & 63.68 & \textbf{60.71}  & 64.27  & \textbf{55.66}  \\ 
SemAtt  & \textbf{92.5} & 92.4  & 95.80 & \textbf{68.17} & 68.37 & \textbf{57.45}  \\
\hline
dBERT($\alpha$) & 0 & 0.45 & 0 & 0.45 & 0 & 0.45\\ \hline
TF  & \textbf{91.7} & 91.1   & 99.78 & \textbf{98.02} & 59.12  & \textbf{53.30}   \\
BAE  & \textbf{91.7} & 91.1 & 68.70 & \textbf{63.45}  & 61.37  & \textbf{54.21}  \\ 
SemAtt  & \textbf{91.7} & 91.1  & 99.02 & \textbf{82.15} & 67.01 & \textbf{57.37}  \\
\bottomrule
\end{tabular}
\vspace{-1em}
\caption{Comparison of standard models and LS models for various attacks on OOD data where $\alpha$ denotes the label smoothing factor ($\alpha$=0: no LS).  }
\label{ood_result}

\end{table}

In Table \ref{ood_result}, we again see that label-smoothing helps produce more robust models in the OOD setting although with less gain compared to iD setting. This is a challenging setting, as evidenced by the significant performance drop in the clean accuracy as compared to the in-domain setting. We also see that the standard models make over-confident errors on successfully attacked adversarial examples, when compared to label-smoothed models. 

\subsection{Qualitative Results}

In this section, we try to understand how the generated adversarial examples differ for label smoothed and standard models. First we look at some qualitative examples: in Table \ref{tab:adv_samples}, we show some examples (clean text) for which the different attack schemes fails to craft an attack for the label smoothed model but successfully attacks the  standard model.
\begin{table}[!htb]
\footnotesize
\vspace{-0.5em}
\begin{tabular}{p{1.7em}p{1.7em}p{1.7em} p{13.5em}}
\toprule
Victim & Attack & \multicolumn{2}{l}{Text} \\ \hline
 SST2 & BAE & clean & at once half-baked and overheated. \\
 BERT & & adv &at once \textcolor{red}{warm} and overheated .  \\ \hline
 MR & TF & clean & no surprises .\\
 dBERT& & adv  & no \textcolor{red}{surprise} . \\
\toprule
\end{tabular}
\caption{Examples for which an attack could be found for the standard model but not for the label smoothed model. The Victim column  shows the dataset and the pretrained model (dBERT denotes distilBERT). }
\vspace{-1em}
\label{tab:adv_samples}
\end{table}

We also performed automatic evaluation of the quality of the adversarial examples for standard and label smoothed models, adopting standard metrics from previous studies \cite{textfooler, clare}. Ideally, we want the adversarial sentences to be free of grammar errors, fluent, and semantically similar to the clean text. This can be quantified using metrics such as grammar errors, perplexity, and similarity scores (compared to the clean text). The reported scores for each metric are computed over only the successful adversarial examples, for each attack and model type.\footnote{Additional details can be found in Appendix\ref{attack_evaluation}.}

\begin{table}[!htb]
\footnotesize
\centering
\begin{tabular}{P{4em} P{2em} P{2em}P{1.8em} P{1.8em}P{1.5em}P{1.5em}} \toprule
\textbf{SST-2} & \multicolumn{2}{c}{\begin{tabular}{@{}c@{}}Perplexity ($\uparrow$) \end{tabular}}  & \multicolumn{2}{c}{\begin{tabular}{@{}c@{}}Similarity \\ Score ($\downarrow$) \end{tabular}} &\multicolumn{2}{c}{\begin{tabular}{@{}c@{}}Grammar \\ Error ($\uparrow$) \end{tabular}} \\ \hline
BERT ($\alpha$) & 0 & 0.45 & 0 & 0.45 & 0 & 0.45\\ \hline
TF  & 400.31 & \textbf{447.58} &  0.800 & \textbf{0.779} &  0.33 & \textbf{0.38} \\
BAE  &  300.74 & \textbf{305.28} & 0.867 & \textbf{0.855}  & $-$0.05 &  $-$\textbf{0.04}    \\ \toprule

\textbf{AG\_News}  & \multicolumn{2}{c}{\begin{tabular}{@{}c@{}}Perplexity ($\uparrow$) \end{tabular}}  & \multicolumn{2}{c}{\begin{tabular}{@{}c@{}}Similarity \\ Score ($\downarrow$) \end{tabular}} &\multicolumn{2}{c}{\begin{tabular}{@{}c@{}}Grammar \\ Error ($\uparrow$) \end{tabular}} \\ \hline
BERT ($\alpha$) & 0 & 0.45 & 0 & 0.45 & 0 & 0.45\\ \hline
TF  & 342.02 & \textbf{355.87} &  0.782 & \textbf{0.772} & 1.37 & \textbf{1.40}   \\
BAE  & 169.37 & \textbf{170.73} &  0.851 & \textbf{0.845} & 0.97 & \textbf{1.00}\\ \bottomrule

\end{tabular}
\caption{Evaluation of adversarial text examples. The results in bold indicates worse adversarial attack quality. }
\vspace{-1em}
\label{adveval}
\end{table}

Table \ref{adveval} shows that the quality of generated adversarial examples on label smoothed models is worse than those on standard models for different metrics, suggesting that the adversarial sentences generated by standard models are easier to perceive. This further demonstrates that label smoothing makes it harder to find adversarial vulnerabilities.


\section{Conclusion}
We presented an extensive empirical study to investigate the effect of label smoothing techniques on adversarial robustness for various NLP tasks, for various victim models and adversarial attacks. Our results demonstrate that label smoothing imparts implicit robustness to models, even under domain shifts. This first work on the effects of LS for text adversarial attacks, complemented with prior work on LS and implicit calibration \cite{desai2020calibration,DanRo21}, is an important step towards developing robust, reliable models. In the future, it would be interesting to explore the combination of label smoothing with other regularization and adversarial training techniques to further enhance the adversarial robustness of NLP models.


\section{Limitations}

One limitation of our work is that we focus on robustness of pre-trained transformer language models against word-level adversarial attacks, which is the most common setting in this area. Future work could extend this empirical study to other types of attacks (for example, character-level and sentence-level attacks) and for diverse types of architectures. Further, it will be very interesting to theoretically understand how label smoothing provides (1) the implicit robustness to text adversarial attacks and (2) mitigates over-confident predictions on the adversarially attacked examples.
\section{Ethics Statement}
Adversarial examples present a severe risk to machine learning systems, especially when deployed in real-world risk sensitive applications. With the ubiquity of textual information in real-world applications, it is extremely important to defend against adversarial examples and also to understand the robustness properties of commonly used techniques like Label Smoothing. From a societal perspective, by studying the effect of this popular regularization strategy, this work empirically shows that it helps robustness against adversarial examples in in-domain and out-of-domain scenarios, for both white-box and black-box attacks across diverse tasks and models. From an ecological perspective, label smoothing does not incur any additional computational cost over standard fine-tuning emphasizing its efficacy as a general-purpose tool to improve calibration and robustness.
\section*{Acknowledgements}
Research was sponsored by the Army Research Office and was accomplished under Grant Number W911NF-20-1-0080. This work was supported by Contract FA8750-19-2-0201 with the US Defense Advanced Research Projects Agency (DARPA). The views expressed are those of the authors and do not reflect the official policy or position of the Department of Defense, the Army Research Office or the U.S. Government. This research was also supported by a gift from AWS AI for research in Trustworthy AI.
\bibliography{anthology,custom}
\bibliographystyle{acl_natbib}

\newpage
\appendix
\section{Appendix}
\begin{itemize}
    \item \textbf{A.1} Pictorial Overview of the Adversarial Attack Framework
    \item \textbf{A.2} Description of the Evaluation Metrics
    \item \textbf{A.3} Details of Automatic Attack Evaluation
    \item \textbf{A.4} Additional results on Movie Review Dataset
    \item \textbf{A.5} Additional white-box attack on label-smoothed models
    \item \textbf{A.6} Additional results for $\alpha=0.1$
    \item \textbf{A.7} Additional results on ALBERT model
    \item \textbf{A.8} Dataset overview and expertiment details
    \item \textbf{A.9} Attack success rate versus label smoothing factors for different attacks (TextFooler and SemAttack)
    \item \textbf{A.10} Average number of word change versus Confidence

\end{itemize}
\subsection{Overview of the Framework}
\begin{figure}[!htb]
\centering
    \includegraphics[width=0.95\linewidth]{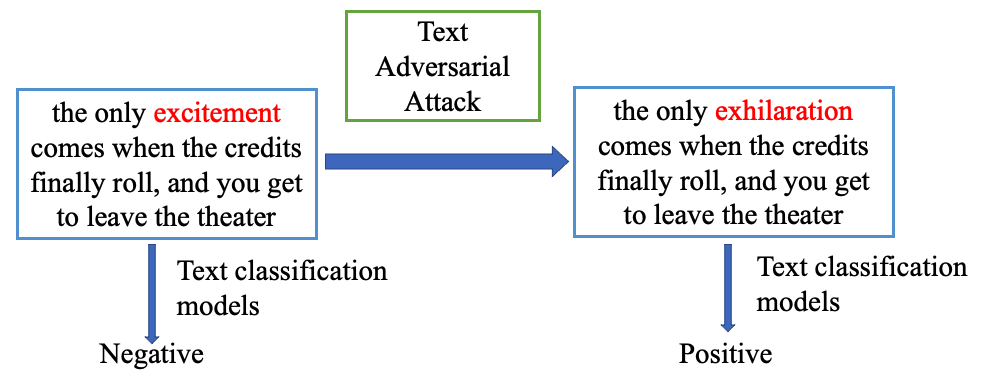}
\caption{Here we show an example generated by word-level adversarial attack TextFooler \citep{textfooler} on SST-2 data. By replacing excitement with its synonym \textit{exhilaration}, the text classification model changes its prediction from Negative to Positive, which is incorrect.  }
\label{fig:adv_text_example}
\end{figure}

\subsection{Evaluation Metrics}
\label{eval_metric}
The followings are the details of evaluation metrics from previous works \citep{queryattack2022,clare}:\\
Clean accuracy = $\frac{\# \text{ of correctly predicted clean examples} }{ \# \text{ of clean examples}}$ \\
Attack Succ. Rate = $\frac{\# \text{ of successful adversarial examples} }{\# \text{ of correctly predicted clean examples}}$ where successful adversarial examples are derived from correctly predicted examples\\
Adv Conf = $\frac{ \text{ sum of confidence of successful adv examples} }{ \# \text{ of successful adversarial examples}}$

\subsection{Attack evaluation}
\label{attack_evaluation}
We performed automatic evaluation of adversarial attacks against standard models and label smoothed models following previous studies \cite{textfooler, clare}. Following are the details of the metrics we used in Table \ref{adveval}: \\
\noindent \textbf{Perplexity} evaluates the fluency of the input using language models. We use GPT-2  \cite{radford2019language} to compute perplexity as in \cite{clare} . \\
\textbf{Similarity Score} determines the similarity between two sentences. We use Sentence Transformers \cite{reimers-2019-sentence-bert} to compute sentence embeddings and then calculate cosine similarity score between the clean examples and the corresponding adversarially modified examples.\\
\textbf{Grammar Error} The average grammar error increments between clean examples and the corresponding adversarially modified example.\footnote{we use \url{https://pypi.org/project/language-tool-python/} to compute grammar error.}

\subsection{Additional results on Movie Review Dataset}
Here we provide results of movie review datasets \citep{mrdataset} under in-domain setting. 
\begin{table}[!htb]
\small
\begin{tabular}{P{0.2\columnwidth} P{0.07\columnwidth}P{0.07\columnwidth} P{0.07\columnwidth} P{0.07\columnwidth}P{0.05\columnwidth} P{0.06\columnwidth}}
\toprule
\centering
\textbf{MR} & \multicolumn{2}{c}{\begin{tabular}{@{}c@{}}Clean \\ Acc ($\uparrow$) \end{tabular}}  & \multicolumn{2}{c}{\begin{tabular}{@{}c@{}} Attack Success \\ Rate ($\downarrow$) \end{tabular}} &\multicolumn{2}{c}{\begin{tabular}{@{}c@{}}Adv \\ Conf ($\downarrow$) \end{tabular}} \\ \hline
BERT ($\alpha$) & 0 & 0.45 & 0 & 0.45 & 0 & 0.45\\ \hline
TextFooler  & \textbf{84.4} & 83.7 & 92.54 & \textbf{92.0} & 67.93 & \textbf{58.33}  \\
BAE & \textbf{84.4} & 83.7 & 62.09 & \textbf{61.17} & 74.33 & \textbf{62.4} \\ 
SemAtt & \textbf{84.4} & 83.7 & 83.18 & \textbf{76.34} & 68.8 & \textbf{58.18} \\ \hline
distilBERT($\alpha$) & 0 & 0.45 & 0 & 0.45 & 0 & 0.45\\ \hline
TextFooler  & 82.3 & \textbf{82.6} & \textbf{94.9} & 95.88 & 64.64 & \textbf{57.17}  \\
BAE & 82.3 & \textbf{82.6} & 67.31 & \textbf{67.19} & 70.54 & \textbf{60.88}\\ 
SemAtt & 82.3 & \textbf{82.6} & 90.16 & \textbf{87.77} & 65.55 & \textbf{57.33}  \\ \bottomrule
\end{tabular}
\caption{Comparison of standard models and label smoothed models against various attacks for Movie Review dataset.}
\end{table}

\subsection{Additional results on an additional white-box attack}
In this section, we use another recent popular white-box attack named Gradient-based Attack \cite{guo2021gradientbased}. This is a gradient-based approach that searches for a parameterized word-level adversarial attack distribution, and then samples adversarial examples from the distribution. We run this attack on standard and label smoothed BERT models and the results are listed below. 
\begin{table}[!htb]
\small
\begin{tabular}{P{0.23\columnwidth} P{0.06\columnwidth}P{0.06\columnwidth}P{0.05\columnwidth} P{0.05\columnwidth}P{0.05\columnwidth}P{0.06\columnwidth}}
\toprule
\centering
\begin{tabular}{@{}c@{}}\textbf{Grad} \\ \textbf{Attack} \end{tabular} & \multicolumn{2}{c}{\begin{tabular}{@{}c@{}}Clean \\ Acc ($\uparrow$) \end{tabular}}  & \multicolumn{2}{c}{\begin{tabular}{@{}c@{}} Attack Succ\\ Rate($\downarrow$) \end{tabular}} &\multicolumn{2}{c}{\begin{tabular}{@{}c@{}}Adv\\ Conf ($\downarrow$) \end{tabular}} \\ \hline
BERT ($\alpha$) & 0 & 0.45 & 0 & 0.45 & 0 & 0.45 \\ \hline
SST-2  & 91.97 & \textbf{92.09} & 98.38 & \textbf{82.94} & 98.75 & \textbf{76.35}  \\
AG\_news & \textbf{94.9} &  94.8 & 98.63 & \textbf{68.88} & 95.35 & \textbf{63.25}\\ 
Yelp & 95.3 &  \textbf{95.5} & 99.90 & \textbf{87.02} & 99.24 & \textbf{76.52} \\
SNLI & 89.7 &  \textbf{90.2} & 96.1 & \textbf{86.36} & 59.99 & \textbf{37.28}\\ 
SST2 $\rightarrow$ Yelp & \textbf{88.6} &  88.4 & 99.89 & \textbf{94.84} & 98.37 & \textbf{77.52}\\ 
 \bottomrule
\end{tabular}
\caption{Comparison of standard models and label smoothed BERT models against gradient-based attack across different datasets.}
\end{table}

We observe that the label smoothing also help with adversarial robustness against this attack across four datasets under iD setting. The results also show that, similar to SemAttack, the grad-based attack benefits more from label smoothing compared to black-box attacks like TextFooler and BAE. 

\subsection{Additional results of $\alpha = 0.1$}

Table \ref{main-0.1} and \ref{ood_result_0.1} are the additional results to show when label smoothing $\alpha = 0.1$, how the adversarial robustness of fine-tuned language models changes under iD and OOD scenarios.

 \begin{table}[!htb]
\small
\begin{tabular}{P{5em} P{1.5em} P{1.5em}P{2.5em} P{2.5em}P{1.5em}P{1.5em}} \toprule
\textbf{SST-2}   & \multicolumn{2}{c}{\begin{tabular}{@{}c@{}}Clean \\ Acc ($\uparrow$) \end{tabular}}  & \multicolumn{2}{c}{\begin{tabular}{@{}c@{}} Attack Success \\ Rate ($\downarrow$) \end{tabular}} &\multicolumn{2}{c}{\begin{tabular}{@{}c@{}}Adv \\ Conf ($\downarrow$) \end{tabular}} \\ \hline
BERT ($\alpha$) & 0 & 0.1 & 0 & 0.1 & 0 & 0.1\\ \hline
TF  & 91.97 & \textbf{92.2}  & 96.38 & \textbf{94.4} & 78.43 & \textbf{74.39}  \\
BAE  & 91.97 & \textbf{92.2} & 57.11 & \textbf{55.22}  & 86.92  & \textbf{82.29}  \\  \hline
distilBERT($\alpha$) & 0 & 0.1 & 0 & 0.1 & 0 & 0.1\\ \hline
TF  & 89.56 &  \textbf{89.68}  & 96.29 & \textbf{95.14} & 76.28 & \textbf{70.77} \\
BAE  & 89.56 &  \textbf{89.68} & 59.28  & \textbf{58.44} & 83.55 & \textbf{78.16}   \\ 
\end{tabular}
\begin{tabular}{P{5em} P{1.5em} P{1.5em}P{2.5em} P{2.5em}P{1.5em}P{1.5em}} \toprule
\textbf{AG\_news}   & \multicolumn{2}{c}{\begin{tabular}{@{}c@{}}Clean \\ Acc ($\uparrow$) \end{tabular}}  & \multicolumn{2}{c}{\begin{tabular}{@{}c@{}} Attack Success \\ Rate ($\downarrow$) \end{tabular}} &\multicolumn{2}{c}{\begin{tabular}{@{}c@{}}Adv \\ Conf ($\downarrow$) \end{tabular}} \\ \hline
BERT ($\alpha$) & 0 & 0.1 & 0 & 0.1 & 0 & 0.1\\ \hline
TF & 94.83 & \textbf{95.0} & 88.26 & \textbf{78.39} & 59.02 & \textbf{55.17} \\
BAE  & 94.83 & \textbf{95.0} & 74.83 & \textbf{65.58} & 60.66 & \textbf{56.24}\\ \hline
distilBERT($\alpha$) & 0 & 0.1 & 0 & 0.1 & 0 & 0.1\\ \hline
TF  & \textbf{94.73} & 94.53 & 90.11 & \textbf{81.66} & 57.6 &  \textbf{53.43}    \\
BAE & \textbf{94.73} & 94.53    & 74.83 & \textbf{67.7} & 60.01 & \textbf{54.64}
\end{tabular}
\begin{tabular}{P{5em} P{1.5em} P{1.5em}P{2.5em} P{2.5em}P{1.5em}P{1.5em}} \toprule
\textbf{Yelp}& \multicolumn{2}{c}{\begin{tabular}{@{}c@{}}Clean \\ Acc ($\uparrow$) \end{tabular}}  & \multicolumn{2}{c}{\begin{tabular}{@{}c@{}} Attack Success \\ Rate ($\downarrow$) \end{tabular}} &\multicolumn{2}{c}{\begin{tabular}{@{}c@{}}Adv \\ Conf ($\downarrow$) \end{tabular}} \\ \hline
BERT ($\alpha$) & 0 & 0.1 & 0 & 0.1 & 0 & 0.1\\ \hline
TF  & 97.73 & \textbf{97.77} & 99.32 & \textbf{97.99}  & 64.85  & \textbf{63.18} \\
BAE  & 97.73 & \textbf{97.77} & 55.35 & \textbf{52.88}  &  68.28  & \textbf{66.28}   \\  \hline
distilBERT($\alpha$) & 0 & 0.1 & 0 & 0.1 & 0 & 0.1\\ \hline
TF & 97.47 & \textbf{97.5} & 99.45 & \textbf{98.91} & 61.75 & \textbf{60.35} \\
BAE  & 97.47 & \textbf{97.5} & 58.14 & \textbf{51.86} & 64.27 & \textbf{63.04} \\ 
\bottomrule
\end{tabular}
\begin{tabular}{P{5em} P{1.5em} P{1.5em}P{2.5em} P{2.5em}P{1.5em}P{1.5em}} 

\textbf{SNLI}  & \multicolumn{2}{c}{\begin{tabular}{@{}c@{}}Clean \\ Acc ($\uparrow$) \end{tabular}}  & \multicolumn{2}{c}{\begin{tabular}{@{}c@{}} Attack Success \\ Rate ($\downarrow$) \end{tabular}} &\multicolumn{2}{c}{\begin{tabular}{@{}c@{}}Adv \\ Conf ($\downarrow$) \end{tabular}} \\ \hline
BERT ($\alpha$) & 0 & 0.1 & 0 & 0.1 & 0 & 0.1\\ \hline
TF & \textbf{89.56} & 88.87 & \textbf{96.5} & 96.74 & 68.83 & \textbf{64.96}\\
BAE & \textbf{89.56} & 88.87 & \textbf{74.95} & 75.1 & 76.13 & \textbf{72.65} \\ \hline
distilBERT($\alpha$) & 0 & 0.1 & 0 & 0.1 & 0 & 0.1\\ \hline
TF  & \textbf{87.27} & 87.03 & 98.12 & \textbf{96.94} & 65.19 & \textbf{62.41} \\
BAE  & \textbf{87.27} & 87.03 & 74.08 & \textbf{73.82} & 72.89  & \textbf{69.57}  \\ 
\bottomrule
\end{tabular}
\caption{Comparison of standard models and label smoothed models against various attacks for in-domain data  where $\alpha$ denotes the label smoothing factor, 0 indicating no LS. \footnote{Adversarial confidence is computed on successful adversarial examples only.} $\uparrow$ ($\downarrow$) denotes higher (lower) is better respectively. }
\label{main-0.1}
\end{table}

\begin{table}[!htb]
\small
\centering
\begin{tabular}{P{5em} P{1.5em} P{1.5em}P{2.5em} P{2.5em}P{1.5em}P{1.5em}} \toprule
{\begin{tabular}{@{}c@{}}\textbf{SNLI} $\rightarrow$ \textbf{MNLI} \end{tabular}}  & \multicolumn{2}{c}{\begin{tabular}{@{}c@{}}Clean \\ Acc ($\uparrow$) \end{tabular}}  & \multicolumn{2}{c}{\begin{tabular}{@{}c@{}} Attack Success \\ Rate ($\downarrow$) \end{tabular}} &\multicolumn{2}{c}{\begin{tabular}{@{}c@{}}Adv \\ Conf ($\downarrow$) \end{tabular}} \\ \hline
BERT ($\alpha$) & 0 & 0.1 & 0 & 0.1 & 0 & 0.1 \\ \hline
TextFooler  & \textbf{73.4} & 71.9  & \textbf{94.82} & 94.85  & 58.04 & \textbf{48.56}   \\
BAE  & \textbf{73.4} & 71.9  & 82.56 &  \textbf{77.19} & 63  & \textbf{49.3}  \\ 
\hline
distilBERT($\alpha$) & 0 & 0.45 & 0 & 0.45 & 0 & 0.45\\ \hline
TextFooler  & \textbf{65.4} & 65.2 & 94.5 & \textbf{94.17}    & 54.54 & \textbf{52.63}  \\
BAE  & \textbf{65.4} & 65.2 & 77.68 & \textbf{75.15}  & 58.88  & \textbf{56.16}  \\ 
\end{tabular}
\begin{tabular}{P{5em} P{1.5em} P{1.5em}P{2.5em} P{2.5em}P{1.5em}P{1.5em}} \toprule
{\begin{tabular}{@{}c@{}}\textbf{SST-2} $\rightarrow$ \textbf{Yelp} \end{tabular}}  & \multicolumn{2}{c}{\begin{tabular}{@{}c@{}}Clean \\ Acc ($\uparrow$) \end{tabular}}  & \multicolumn{2}{c}{\begin{tabular}{@{}c@{}} Attack Success \\ Rate ($\downarrow$) \end{tabular}} &\multicolumn{2}{c}{\begin{tabular}{@{}c@{}}Adv \\ Conf ($\downarrow$) \end{tabular}} \\ \hline
BERT ($\alpha$) & 0 & 0.1 & 0 & 0.1 & 0 & 0.1\\ \hline
TextFooler  & \textbf{92.5} & 92.0   & 99.57 & \textbf{99.13} & 60.8  & \textbf{58.13}   \\
BAE  & \textbf{92.5} & 92.0 & 63.68 & \textbf{63.37}  & 64.27  & \textbf{60.63}  \\ 
\hline
distilBERT($\alpha$) & 0 & 0.45 & 0 & 0.45 & 0 & 0.45\\ \hline
TextFooler  & \textbf{91.7} & 91.4   & 99.78 & \textbf{99.34} & 59.12  & \textbf{56.42}   \\
BAE  & \textbf{91.7} & 91.4 & 68.7 & \textbf{67.07}  & 61.37  & \textbf{57.73}  \\ 
\bottomrule
\end{tabular}
\caption{Comparison of standard models versus label smoothed models against various attacks for OOD data where $\alpha$ denotes the label smoothing factor ($\alpha$=0: no LS).  $\uparrow$ ($\downarrow$) denotes higher (lower) is better respectively. }
\label{ood_result_0.1}
\end{table}

Table \ref{advlabel_0.1} are the additional results for adversarial label smoothing $\alpha = 0.1$.

\begin{table}[!htb]
\footnotesize
 \singlespacing
\vspace{-1em}
\begin{tabular}{P{3em} P{2em} P{2em}P{2em} P{2em}P{2em}P{2em}} \toprule
\textbf{SNLI}   & \multicolumn{2}{c}{\begin{tabular}{@{}c@{}}Clean \\ Acc ($\uparrow$) \end{tabular}}  & \multicolumn{2}{c}{\begin{tabular}{@{}c@{}} Attack Success\\Rate ($\downarrow$)\end{tabular}} &\multicolumn{2}{c}{\begin{tabular}{@{}c@{}}Adv \\ Conf ($\downarrow$) \end{tabular}} \\ \hline
BERT($\alpha$) & 0 & 0.1 & 0 & 0.1 & 0 & 0.1\\ \hline
TF & 89.56 & \textbf{90.4} & 96.5 & \textbf{95.02} & 68.27 & \textbf{67.54}\\
BAE & 89.56 & \textbf{90.4} & \textbf{74.95} & 75.96 & 76.13 & \textbf{73.83}\\
\bottomrule
\textbf{AG\_news}   & \multicolumn{2}{c}{\begin{tabular}{@{}c@{}}Clean \\ Acc ($\uparrow$) \end{tabular}}  & \multicolumn{2}{c}{\begin{tabular}{@{}c@{}} Attack Success\\Rate ($\downarrow$)\end{tabular}} &\multicolumn{2}{c}{\begin{tabular}{@{}c@{}}Adv \\ Conf ($\downarrow$) \end{tabular}} \\ \hline
BERT($\alpha$) & 0 & 0.1 & 0 & 0.1 & 0 & 0.1\\ \hline
TF & \textbf{94.83} & 94.6 & 88.26 & \textbf{85.27} & 59.02 & \textbf{53.17} \\
BAE  & \textbf{94.83} & 94.6 & 74.83 & \textbf{69.1} & 60.66 & \textbf{54.99} \\  
\bottomrule
\end{tabular}
\caption{Comparison of standard models versus models trained with ALS against various attacks on SNLI and AG\_news.  $\uparrow$ ($\downarrow$) denotes higher (lower) is better respectively.}
\label{advlabel_0.1}

\end{table}

\subsection{Additional results on ALBERT}
In this section, we include experiment results for standard ALBERT and label smoothed ALBERT in Table \ref{albert}. We observe that the label smoothing technique also improves adversarial robustness of ALBERT model across different datasets. 

\begin{table}[H]
\small
\begin{tabular}{P{0.2\columnwidth} P{0.06\columnwidth}P{0.06\columnwidth} P{0.06\columnwidth} P{0.06\columnwidth}P{0.06\columnwidth} P{0.06\columnwidth}}
\toprule
\textbf{SST-2}   & \multicolumn{2}{c}{\begin{tabular}{@{}c@{}}Clean \\ Acc ($\uparrow$) \end{tabular}}  & \multicolumn{2}{c}{\begin{tabular}{@{}c@{}} Attack Success \\ Rate ($\downarrow$) \end{tabular}} &\multicolumn{2}{c}{\begin{tabular}{@{}c@{}} Adv \\ Conf ($\downarrow$) \end{tabular}} \\ \hline
$\alpha$ & 0 & 0.45 & 0 & 0.45 & 0 & 0.45\\ \hline
TF  & 92.66 & \textbf{92.78} & 94.68 & \textbf{90.73} & 76.29 & \textbf{65.63}\\
BAE & 92.66& \textbf{92.78} & \textbf{60.15} & 65.02 & 83.67 & \textbf{70.17}\\
\end{tabular}

\begin{tabular}{P{0.2\columnwidth} P{0.06\columnwidth}P{0.06\columnwidth} P{0.06\columnwidth} P{0.06\columnwidth}P{0.06\columnwidth} P{0.06\columnwidth}}
\toprule
\textbf{AG\_news}   & \multicolumn{2}{c}{\begin{tabular}{@{}c@{}}Clean \\ Acc ($\uparrow$) \end{tabular}}  & \multicolumn{2}{c}{\begin{tabular}{@{}c@{}}Attack Success \\ Rate ($\downarrow$)  \end{tabular}} &\multicolumn{2}{c}{\begin{tabular}{@{}c@{}} Adv \\ Conf ($\downarrow$) \end{tabular}} \\ \hline
$\alpha$ & 0 & 0.45 & 0 & 0.45 & 0 & 0.45\\ \hline
TF  & \textbf{94.9} & 94.5 & 77.66 & \textbf{56.72} & 58.78 & \textbf{42.59}\\
BAE & \textbf{94.9} & 94.5 & 65.54 & \textbf{49.74}  & 59.98 & \textbf{43.79} \\
\end{tabular}

\begin{tabular}{P{0.2\columnwidth} P{0.06\columnwidth}P{0.06\columnwidth} P{0.06\columnwidth} P{0.06\columnwidth}P{0.06\columnwidth} P{0.06\columnwidth}}
\toprule
\textbf{SNLI}   & \multicolumn{2}{c}{\begin{tabular}{@{}c@{}}Clean \\ Acc ($\uparrow$) \end{tabular}}  & \multicolumn{2}{c}{\begin{tabular}{@{}c@{}}Attack Success \\ Rate ($\downarrow$) \end{tabular}} &\multicolumn{2}{c}{\begin{tabular}{@{}c@{}}Adv \\ Conf ($\downarrow$) \end{tabular}} \\ \hline
$\alpha$ & 0 & 0.45 & 0 & 0.45 & 0 & 0.45\\ \hline
TF & 90.1 & \textbf{90.3} & 94.89 & \textbf{93.69} & 69.66 & \textbf{53.67} \\
BAE & 90.1 & \textbf{90.3} & 76.91 & \textbf{75.86} & 75.05 & \textbf{56.42} \\ \bottomrule
\end{tabular}
\caption{Comparison of standard models and label smoothed models against TextFooler and BAE attacks for ALBERT model.}
\label{albert}
\end{table}


\subsection{Dataset Overview and Experiments Details}
\label{exp}
\begin{table}[!htb]
\small
\centering
\begin{tabular}{P{0.15\columnwidth} P{0.25\columnwidth}P{0.20\columnwidth} P{0.18\columnwidth}}
\toprule
\textbf{Dataset} & No. of classes & Train/Test size & Avg. Length \\ \hline
\textbf{MR} & 2 & 8530/1066 & 18.64 \\ 
\textbf{SST-2} & 2 &6.7e4/872 &  17.4 \\ 
\textbf{Yelp} & 2 & 5.6e5/3.8e4 & 132.74 \\ 
\textbf{AG\_news} & 4 &1.2e5 /7600 & 38.68 \\ 
\textbf{SNLI} & 3 & 5.5e5 /1e4 & 22.01 \\ 
\textbf{MNLI} & 3 & 3.9e5/ 9815	 & 28.96 \\ \bottomrule
\end{tabular}
\caption{Summary of datasets}
\end{table}
  We use Huggingface \citep{wolf-etal-2020-transformers} to load the dataset and to fine-tune the pre-trained models. All models are fine-tuned for 3 epochs using AdamW optimizer \cite{loshchilov2017decoupled} and the learning rate starts from $5e-6$. The training and attacking are run on an NVIDIA Quadro RTX 6000 GPU (24GB). For both BAE and Textfooler attack, we use the implementation in TextAttack \citep{morris2020textattack} with the default hyper-parameters (Except for AG\_news, we relax the similarity threshld from 0.93 to 0.7 when using BAE attack). The SemAttack is implemented by \citep{semattack} while the generating contextualized embedding space is modified from \citep{embeddingspace}. The reported numbers are the average performance over $3$ random runs of the experiment for iD setting, and the standard deviation is less than 2\%.

\subsection{Attack success rate versus label smoothing factors}
\label{sec:alpha_sweep_textfooler}
\begin{figure}[!htb]
\centering
    \includegraphics[width=0.7\linewidth]{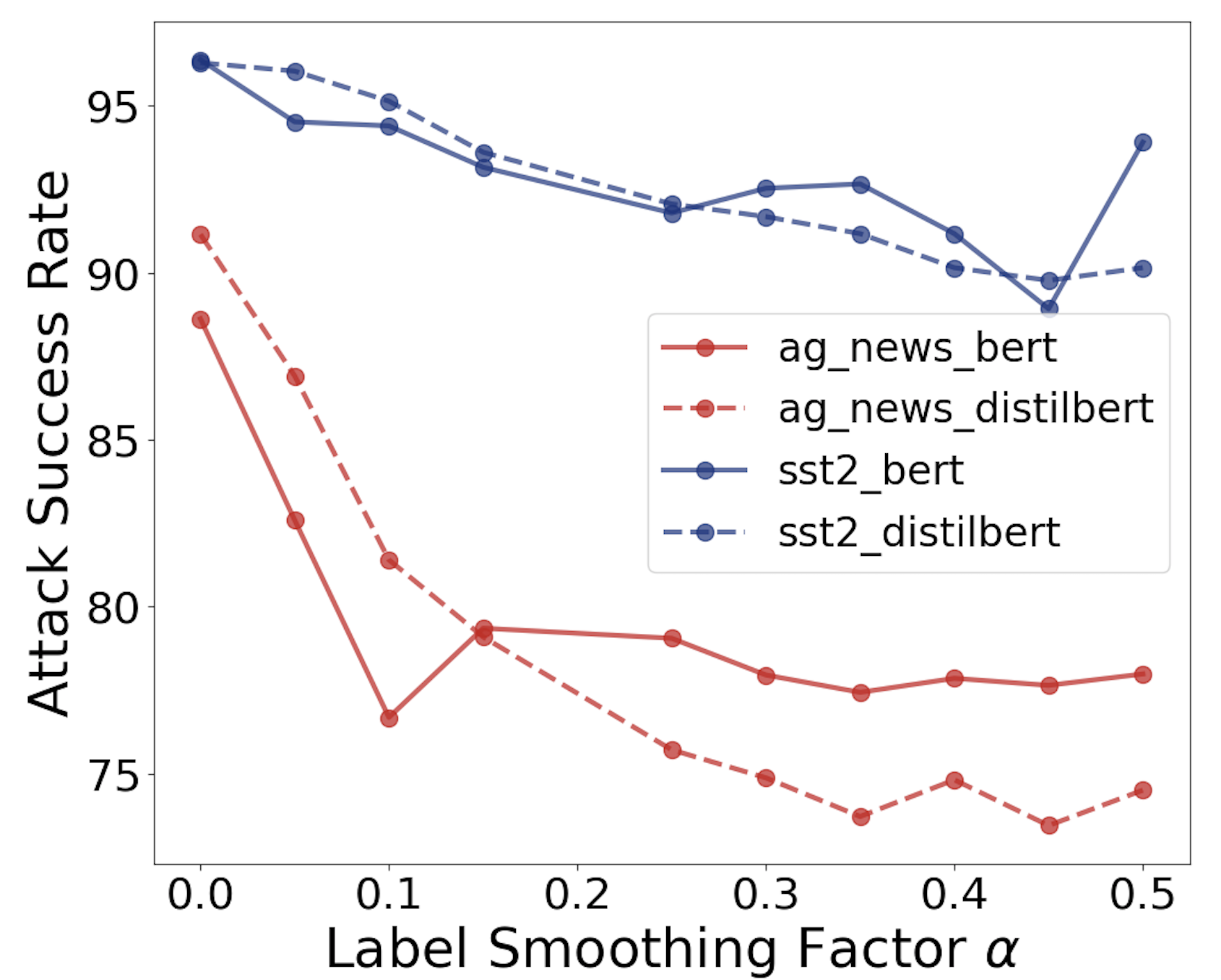}
\caption{Adversarial success rate versus label smoothing factors for the TextFooler attack (on AG News and SST-2.) }
\label{fig:alpha_vs_succ_text}
\end{figure}

\begin{figure}[!htb]
\centering
\includegraphics[width=0.7\linewidth]{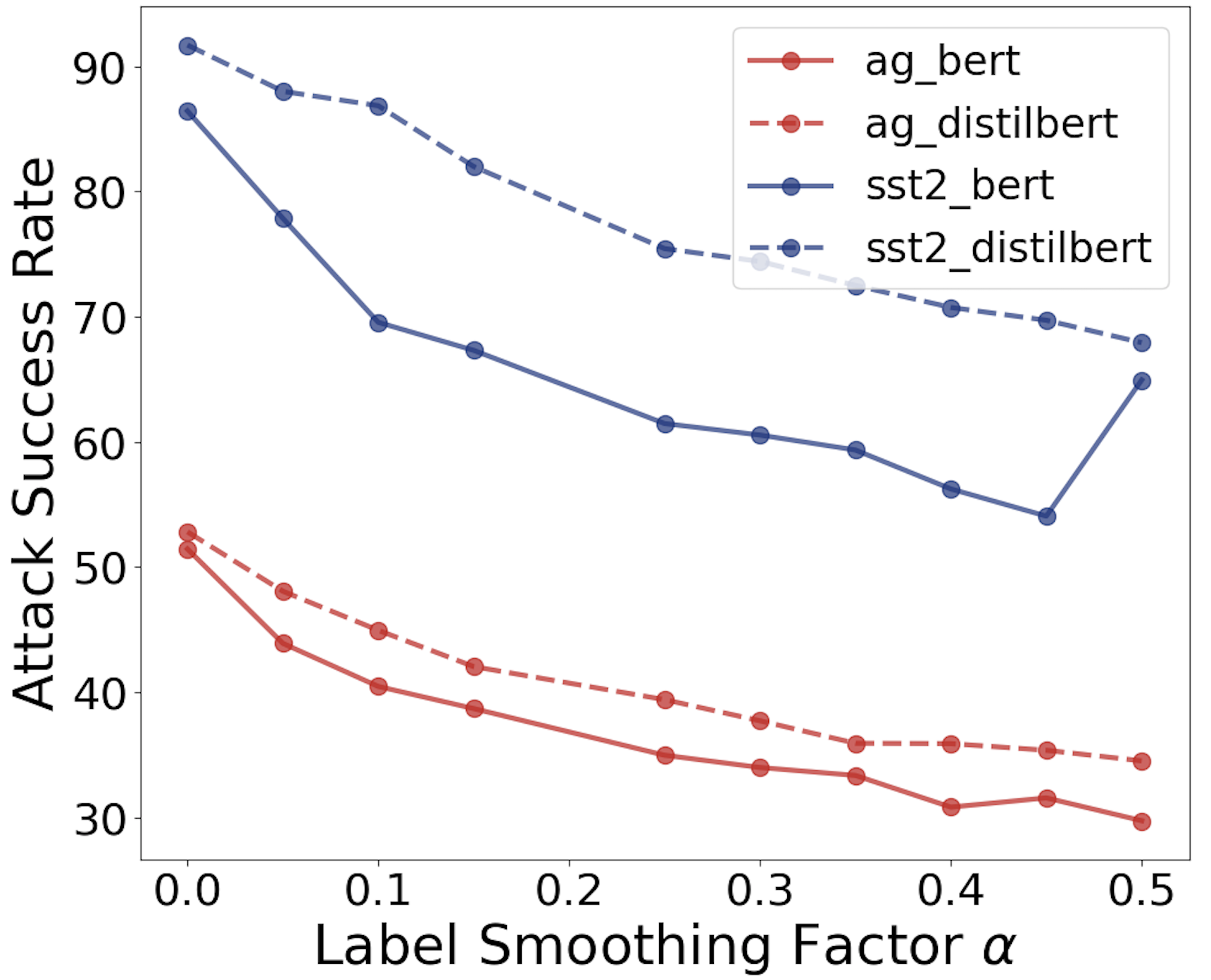}
\caption{Adversarial success rate versus label smoothing factors for the SemAttack (on AG News and SST-2.) }
\label{fig:alpha_vs_succ_sem}
\end{figure}

As mentioned in Section \ref{sec:in_domain}, we plot the attack success rate of BAE attack versus the label smoothing factors. Here, we plot the results for the TextFooler and SemAttack in Figure \ref{fig:alpha_vs_succ_text} and \ref{fig:alpha_vs_succ_sem}, and observe the same tendency as we discussed above.

\begin{figure}[!htb]
\centering
    \includegraphics[width=0.7\linewidth]{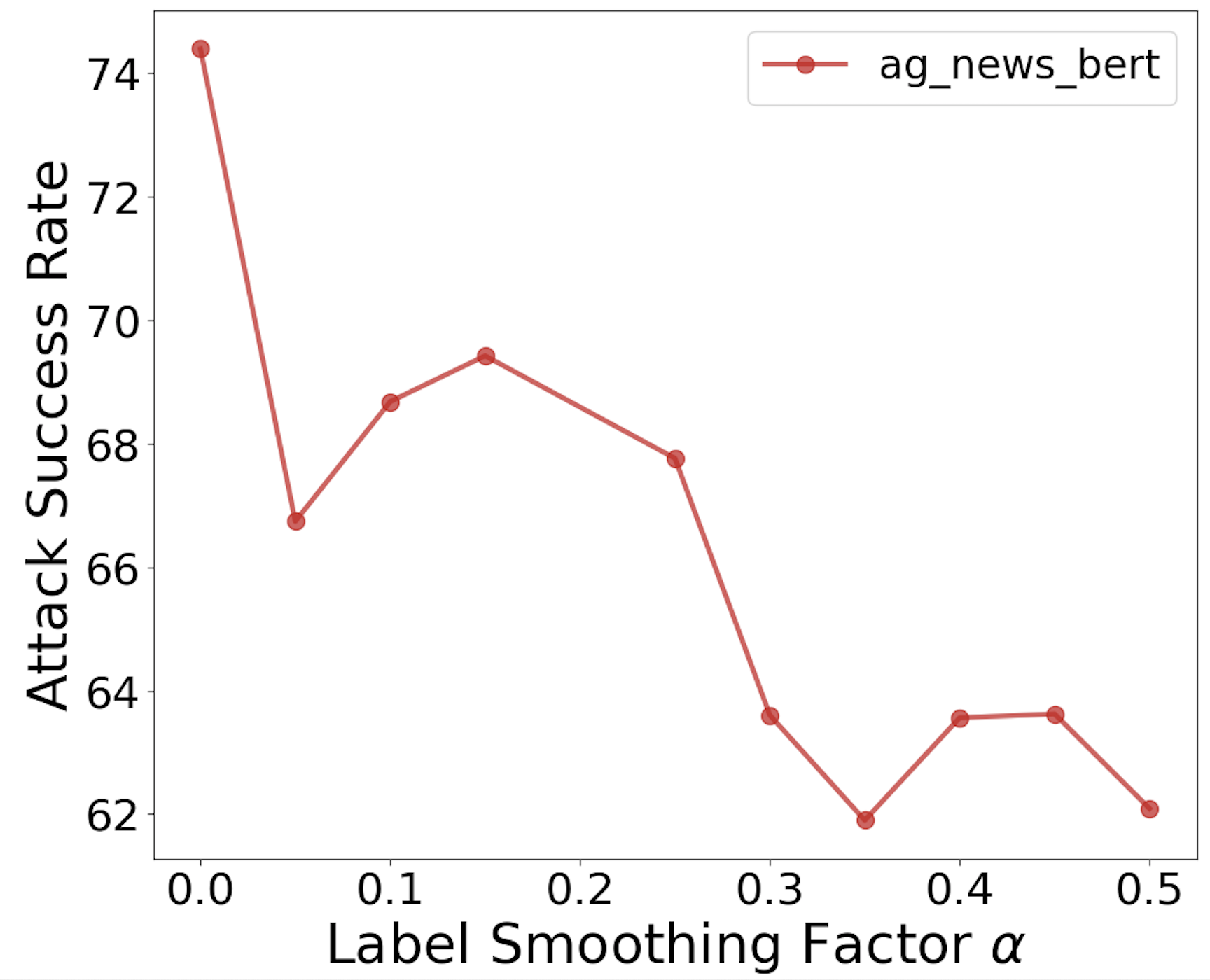}
\caption{Adversarial success rate versus adversarial label smoothing factors for the BAE attack (on AG News). }
\label{fig:alpha_vs_succ_bae_adv}
\end{figure}

\begin{figure}[!htb]
\centering

    \includegraphics[width=0.7\linewidth]{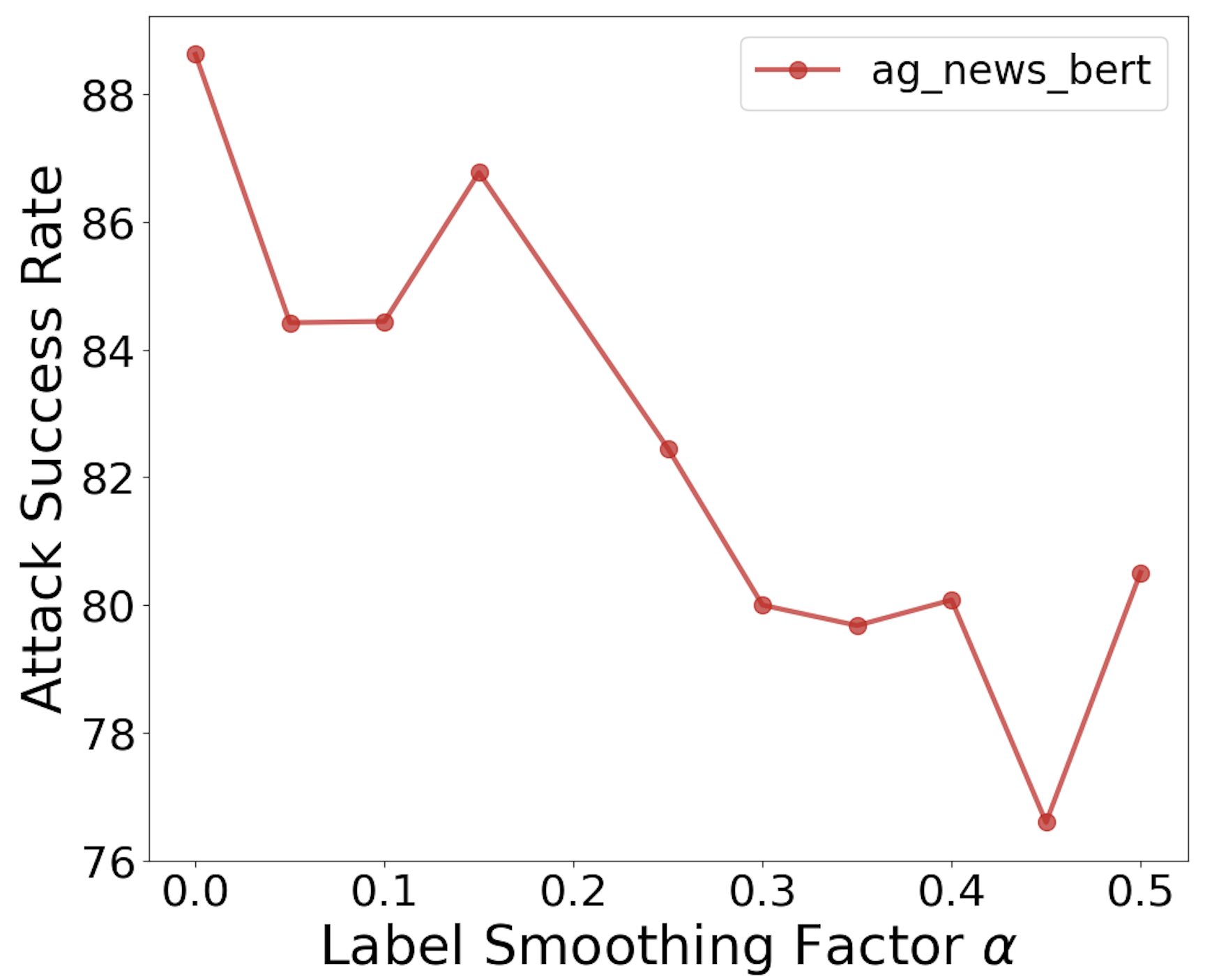}
\caption{Adversarial success rate versus adversarial smoothing factors for the TextFooler attack (on AG News). }
\label{fig:alpha_vs_succ_tf_adv}
\end{figure}

We also plot the attack success rate of BAE/TextFooler attack versus the adversarial label smoothing factors in Figure \ref{fig:alpha_vs_succ_bae_adv} and \ref{fig:alpha_vs_succ_tf_adv}. 

\begin{figure}[!htb]
\centering

\includegraphics[width=0.65\linewidth]{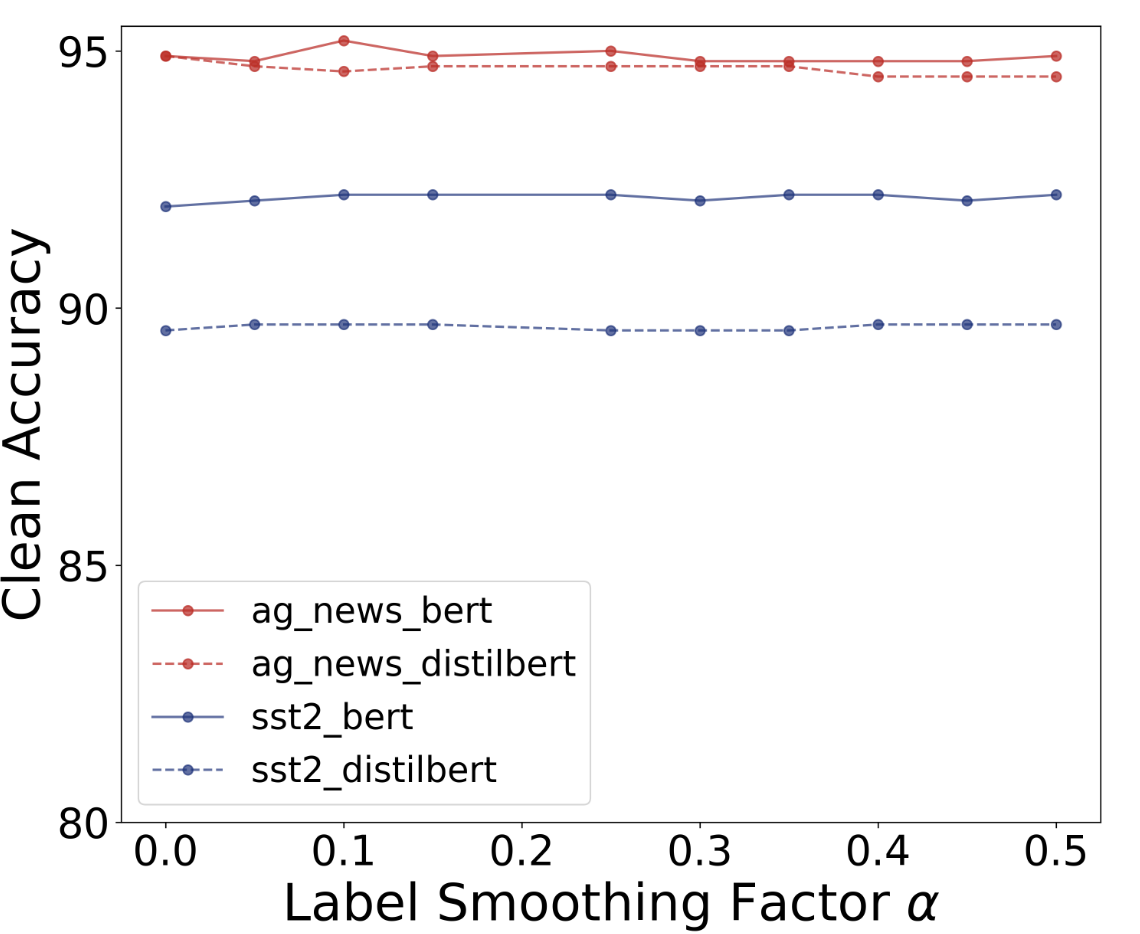}
\caption{Clean accuracy versus label smoothing factors (on AG News and SST-2.) }
\label{fig:alpha_vs_clean}
\end{figure}

We additionally plot the clean accuracy versus the label smoothing factor in Figure \ref{fig:alpha_vs_clean}, and find out that there is not much drop in clean accuracy with increasing the label smoothing factors. 

\subsection{Average number of word change versus Confidence}

Word change rate is defined as the ratio between the number of word replaced after attack and the total number of words in the sentence. Here we plot the bucket-wise word change ratio of adversarial attack versus confidence, and observe that the word change rate for high-confident examples are higher for label smoothed models compared to standard models in most cases. This indicates that it is more difficult to attack label smoothed text classification models. Also note that there is the word change rate is zero because there is no clean texts fall into those two bins. 
\begin{figure}[!htb]
\centering
    \includegraphics[width=0.8\linewidth]{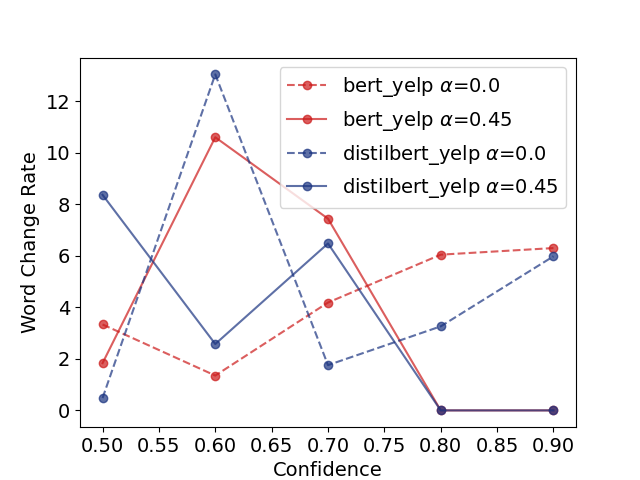}
\caption{Average word change ratio versus confidence for in-domain inputs (No. of buckets: 10 and the number of instances in first 5 buckets [0-0.5] are 0)}
\label{fig:indist-word}
\end{figure}

\begin{figure}[!htb]
\centering
\vspace{-1em}
    \includegraphics[width=0.8\linewidth]{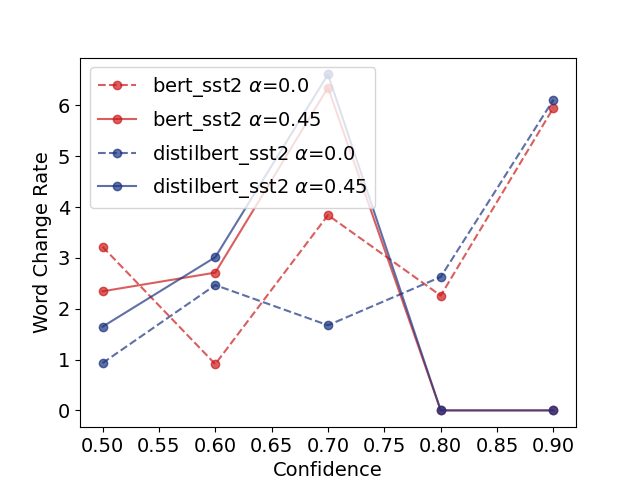}
\caption{Average word change ratio versus confidence for out-of-domain inputs (No. of buckets: 10 and the number of instances in first 5 buckets [0-0.5] are 0)}
\label{fig:outdist-word}
\end{figure}

\begin{figure}[!tb]
\centering
\vspace{-1.25em}
    \includegraphics[width=0.8\linewidth]{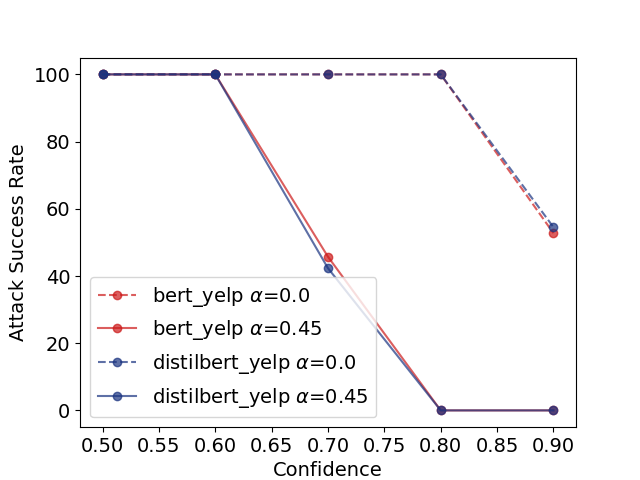}
\caption{Adversarial success rate versus confidence for in-domain (Yelp) inputs. (Number of buckets: 10 and the number of instances in first 5 buckets [0-0.5] are 0).}
\label{fig:indist}
\end{figure}

\begin{figure}[!htb]
\centering
\vspace{-1em}
    \includegraphics[width=0.8\linewidth]{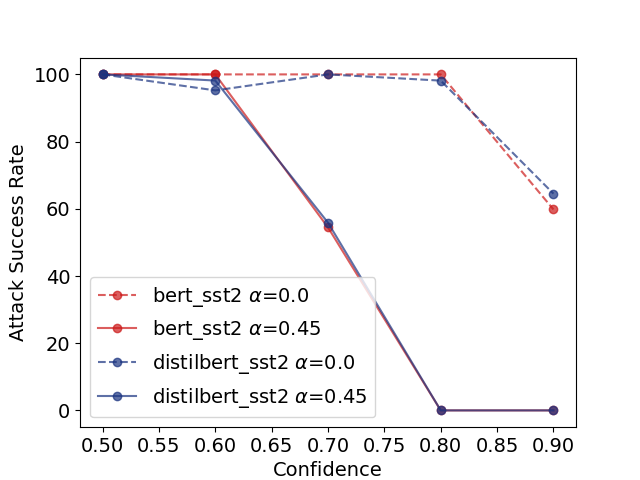}
    \caption{Adversarial success rate versus confidence for OOD inputs in the SST-2 $\rightarrow$ Yelp transfer setting.}
\label{fig:outdist}
\end{figure}

Moreover, we bucket the examples based on the confidence scores, and plot the bucket-wise attack success rate (of the BAE attack on the Yelp dataset) versus confidence in Figure \ref{fig:indist} and Figure \ref{fig:outdist}. We observe that the label smoothing technique improves the adversarial robustness for high confidence score samples significantly.
 In future work, we plan to investigate the variations of robustness in label-smoothed models as a function of the model size.

\label{sec:appendix}

\end{document}